\documentclass[sigconf]{acmart}
\renewcommand\footnotetextcopyrightpermission[1]{} 
\AtBeginDocument{%
  \providecommand\BibTeX{{%
    \normalfont B\kern-0.5em{\scshape i\kern-0.25em b}\kern-0.8em\TeX}}}

\setcopyright{none}
\copyrightyear{}
\acmYear{}
\acmDOI{}
\acmConference[Document Intelligence Workshop at KDD]{The Second Document Intelligence Workshop at KDD}{2021}{Virtual Event}
\acmPrice{}
\acmISBN{}

\begin{document}

\title{The Law of Large Documents: Understanding the Structure of Legal Contracts Using Visual Cues}

\author{Allison Hegel}
\affiliation{%
  \institution{Lexion}
  \city{Seattle, Washington}
  \country{USA}
}
\email{allison@lexion.ai}

\author{Marina Shah}
\affiliation{%
  \institution{Lexion}
  \city{Seattle, Washington}
  \country{USA}
}
\email{marina@lexion.ai}

\author{Genevieve Peaslee}
\affiliation{%
  \institution{Lexion}
  \city{Seattle, Washington}
  \country{USA}
}
\email{genevieve@lexion.ai}

\author{Brendan Roof}
\affiliation{%
  \institution{Lexion}
  \city{Seattle, Washington}
  \country{USA}
}
\email{brendan@lexion.ai}

\author{Emad Elwany}
\affiliation{%
  \institution{Lexion}
  \city{Seattle, Washington}
  \country{USA}
}
\email{emad@lexion.ai}

\renewcommand{\shortauthors}{Hegel, et al.}

\begin{abstract}
Large, pre-trained transformer models like BERT have achieved state-of-the-art results on document understanding tasks, but most implementations can only consider 512 tokens at a time. For many real-world applications, documents can be much longer, and the segmentation strategies typically used on longer documents miss out on document structure and contextual information, hurting their results on downstream tasks. In our work on legal agreements, we find that visual cues such as layout, style, and placement of text in a document are strong features that are crucial to achieving an acceptable level of accuracy on long documents. We measure the impact of incorporating such visual cues, obtained via computer vision methods, on the accuracy of document understanding tasks including document segmentation, entity extraction, and attribute classification. Our method of segmenting documents based on structural metadata out-performs existing methods on four long-document understanding tasks as measured on the Contract Understanding Atticus Dataset.

\end{abstract}



\maketitle

\def\sectionautorefname{Section}
\def\subsectionautorefname{Subsection}

\section{Introduction}

Businesses interact through contracts, and legal departments in many companies are stretched thin managing requests for information about these contracts. Companies need timely reporting on the date contracts are up for renewal, the length of cancellation periods, how much money they should bill, and other details that are embedded in the contract prose. Finding these answers quickly in a large collection of documents, some of which can be hundreds of pages long, can lead to a lot of frustration (and billable hours). At \href{https://lexion.ai/}{Lexion}, we use natural language processing methods to perform this work automatically, extracting key pieces of information from contracts for lawyers to find at a glance \cite{DBLP:journals/corr/abs-1911-00473}.


\begin{figure}[t]
  \centering
  \includegraphics[width=\linewidth]{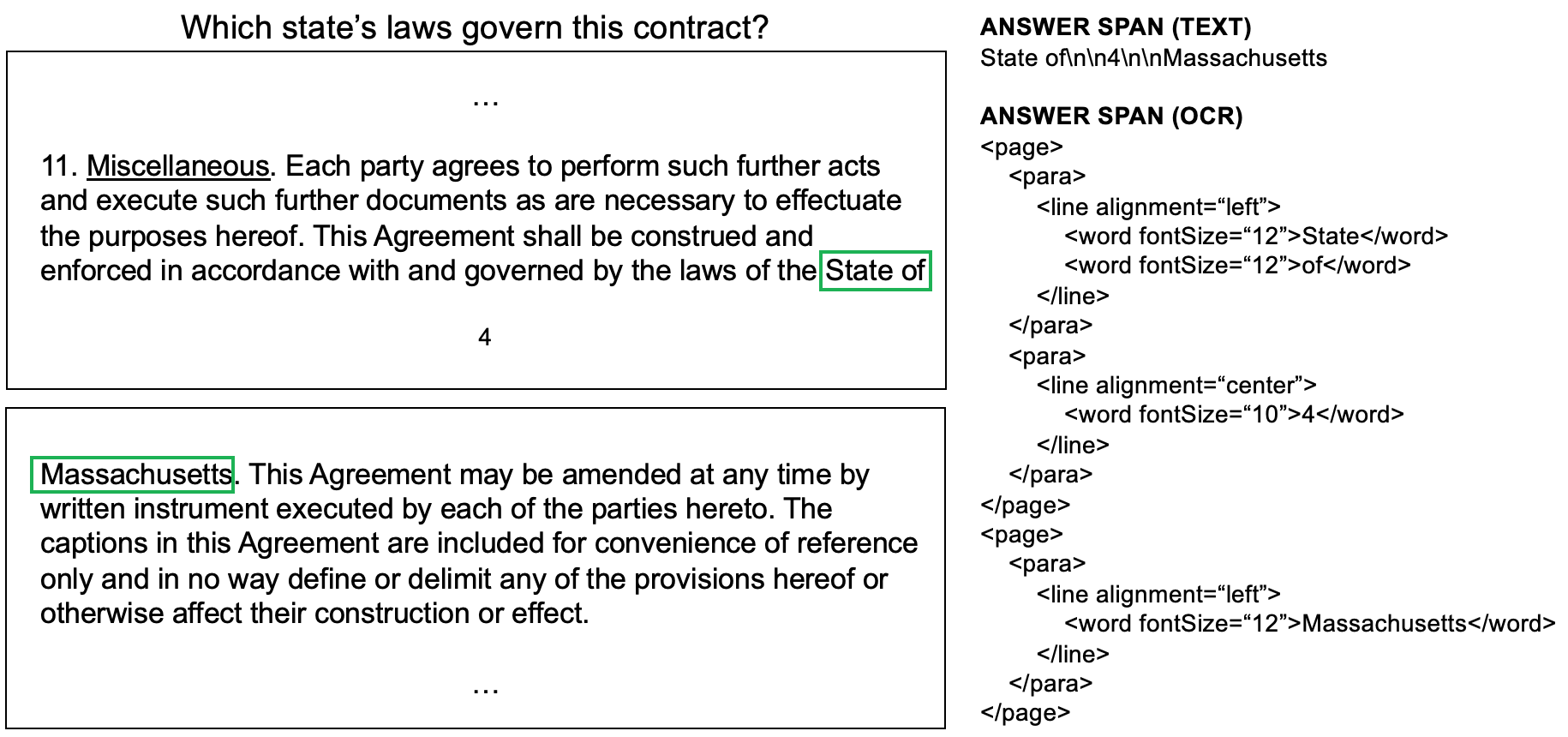}
  \caption{Legal counsel needs to quickly find which state or country's laws will govern an agreement, but this information is often hidden in long documents in sections like ``Miscellaneous''. In this example contract from CUAD \cite{cuad2021}, the desired answer, ``State of Massachusetts'', is broken across a page boundary, with the page number ``4'' in between. To extract this result correctly, a model would need information about the structure and formatting of the document to know that it should discard the page break, newlines, page number, and any other extraneous information common in legal documents such as headers and footers. OCR metadata provides the rich information necessary to find the correct answer span in difficult cases like this one.}
  \label{fig:frontpage}
  \Description{Example of a document understanding task from CUAD.}
\end{figure}

The most powerful recent methods for document understanding are computationally expensive, and they are limited in the length of text they can process at once. To leverage these state-of-the-art model architectures, researchers have developed strategies to segment documents and feed them piecemeal into models. However, these strategies typically discard the document-level context that is essential for complex natural language tasks on long documents. We develop a method that makes use of state-of-the-art models while also maintaining information about the document's structure and formatting. Key to this method is the use of features from Optical Character Recognition (OCR) above and beyond the document's raw text (\autoref{fig:frontpage}).

We evaluate our method on a publicly-available dataset of commercial legal contracts (\autoref{dataset}). Our method (\autoref{experimental_setup}) out-performs prior work on the dataset for four key contract understanding tasks (\autoref{experimental_results}). We also quantify the effect of the OCR features and measure the impact of document length on the difficulty of the task (\autoref{analysis}).

\section{Related Work}

\subsection{Long Document Understanding}

When working with long documents, many models are limited in the length of text they can consider at once. To overcome this limitation, the simplest method is to keep only the portion of the document that fits and discard the rest, but many legal documents contain important information that can be located anywhere in the document. Another method is to split the text into equal-sized sections and input them into the model separately, for example using a sliding window over the documents so that each segment fits within the 512-token maximum length of BERT-based models \cite{cuad2021}. However, splitting documents into sections without any contextual information makes downstream tasks much more difficult (see \autoref{impactofdoclength}).

Most work on long documents circumvents the challenge of splitting documents into meaningful sections by using datasets with clearly demarcated sections, such as Wikipedia articles \cite{choi-etal-2017-coarse, chalkidis-etal-2019-large}, but determining section boundaries is difficult on a corpus of documents with inconsistent structure like legal documents. Research has consistently shown better results on document understanding tasks after splitting documents into sections \cite{sotudeh-gharebagh-etal-2020-guir, choi-etal-2017-coarse, gong-etal-2020-recurrent, wan2019}. As we will show, representing document structure is essential to the section splitting process, and therefore to achieving strong results on downstream tasks.

\subsection{OCR}

Converting images into text via OCR is a precursor to many NLP tasks, but modern OCR tools provide much more than just textual information. OCR methods produce rich metadata including page partitioning and placement, font style and size, justification, and level of indentation, which become features that allow NLP models to incorporate the document-level context that is essential for understanding long documents.

However, OCR features can be difficult to integrate into NLP pipelines. Existing strategies include encoding them as special characters in models like BERT \cite{devlin-etal-2019-bert}, creating encodings to represent the spatial location of each character \cite{katti-etal-2018-chargrid}, or using representations particularly suited to structured data such as graphs \cite{DBLP:journals/corr/abs-2009-05158}.

\section{Dataset} \label{dataset}

Legal documents are especially challenging for NLP tasks because of their length. We use the Contract Understanding Atticus Dataset (CUAD), which contains 510 English-language commercial legal contracts from the public domain \cite{cuad2021}. Each contract is labeled by trained annotators for 41 attributes. Some of these attributes are phrases that must be extracted from the contract as written (entity extraction), and other attributes are yes-or-no values that can be answered from the contract text (classification). Documents in CUAD are longer than those in most NLP datasets, which are often composed of sentences \cite{warstadt2018neural, socher-etal-2013-recursive} or short social media posts like tweets limited to 140 characters \cite{zampieri-etal-2019-predicting, sordoni-etal-2015-neural}. Even document-length corpora consist of documents that are much shorter than legal documents: the average length of a Wikipedia article, for example, is 620 words \cite{wiki:size}, compared to CUAD's average length of 9,594 words. Detailed statistics on CUAD are included in \autoref{app:dataset}.

Since CUAD does not provide section labels, we report section splitting results on our proprietary corpus of contracts, evaluated against section labels created by expert annotators.

\section{Experimental Setup} \label{experimental_setup}

Our end-to-end system is outlined in \autoref{fig:steps} below, beginning with a raw PDF contract and ending with a predicted answer for a document understanding task.

\begin{figure}[ht]
  \centering
  \includegraphics[width=\linewidth]{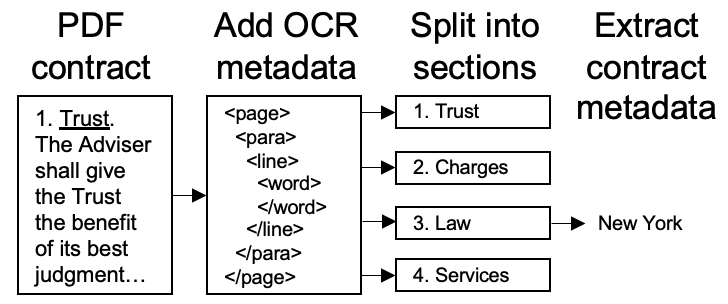}
  \caption{Our end-to-end document understanding system.}
  \label{fig:steps}
  \Description{An outline of the four steps in our document understanding system: starting with a PDF contract, we add OCR metadata, split the contract into sections, and then extract contract metadata.}
\end{figure}

\begin{figure}[ht]
  \centering
  \includegraphics[width=\linewidth]{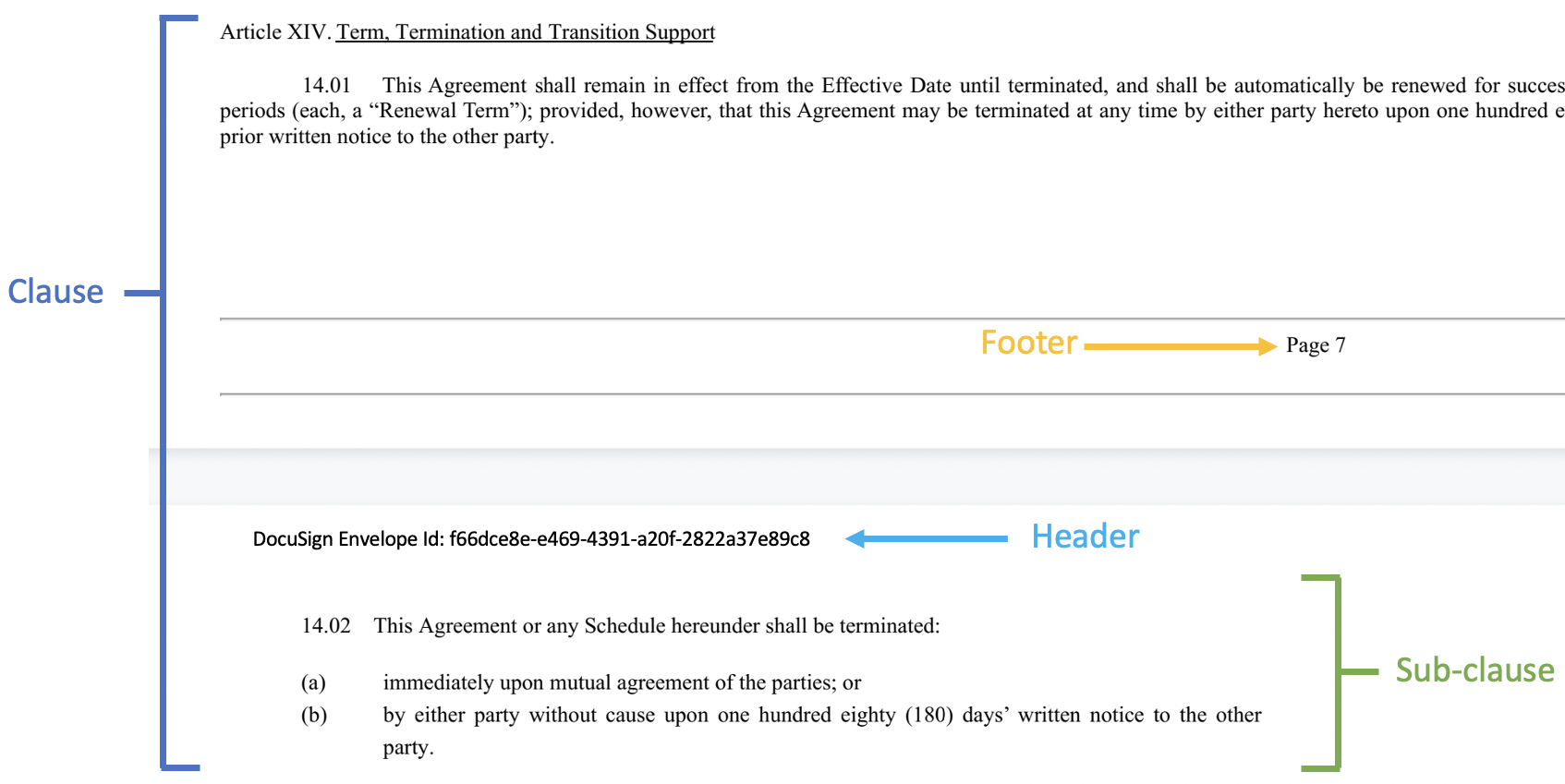}
  \caption{An example document showing each section type.}
  \label{fig:clauses}
  \Description{An example of a legal document where a clause, sub-clause, header, and footer are clearly distinguishable from visual cues.}
\end{figure}

In order to provide focused inputs to our information extraction models, we split documents into coherent sections of several types: clauses, sub-clauses, headers, and footers (\autoref{fig:clauses}). While clauses and sub-clauses contain the information that lawyers are interested in parsing from contracts, it is necessary to detect other types of sections like headers and footers as well because they contain noise that interrupts the flow of text, making it challenging for models to extract the correct values.

We first trained a baseline model for section splitting which takes textual and linguistic features as inputs, with no visual or structural information. This model struggled to accurately identify sections that a human could easily distinguish using visual cues not available to our baseline model.

With the hypothesis that the rich visual data available through OCR would improve performance on section splitting, we tested ablations of our baseline model to measure the impact of four types of visual cues:

\begin{itemize}
\item \textbf{Page layout features}: the proximity of text to various regions of the page (e.g. the beginning, end, center, or margin of a page).
\item \textbf{Text placement features}: the placement of text on the page (e.g. is it centered or aligned).
\item \textbf{Visual grouping features}: which words in the text are grouped together into a paragraph, list, or table.
\item \textbf{Stylistic features}: whether the words are stylized (e.g. bold, italic, or underlined).
\end{itemize}

We find that these visual features improve performance for all section types, with the largest impact on footer detection. We detail these results in \autoref{experimental_results}.

\section{Results} \label{experimental_results}

\subsection{Metrics}

We evaluate our models using precision, recall, and F1 score because many contract attributes suffer from class imbalance. For example, only 15 of the 510 contracts in CUAD (3\%) forbid price changes (the ``Price Restriction'' attribute). If a lawyer wanted to find these 15 contracts, a model that returned none of them would achieve 97\% accuracy, while the recall of such a model would be a more meaningful 0\%.


\subsection{Section Splitting}

To evaluate the impact of visual cues on the accuracy of our section splitting method, we run multiple experiments in which we activate different groups of visual features, and then measure the accuracy on a held out dataset. We compare each group's performance to a baseline model that relies only on text-based linguistic features and low-level layout information, with no access to visual cues from OCR features.

We observe that visual cues have a strong impact, which varies depending on the section type (\autoref{tab:ocr}). At a high level, it is clear that footer detection gains the most from OCR features, but clauses and sub-clauses see significant improvements as well. The results for headers are mixed: recall improves, but at a cost to precision. We look more closely at the reasons for these results in \autoref{impactofocr}.

\begin{table}[!htb]
  \caption{Impact of OCR Features on Section Splitting}
  \label{tab:ocr}
  \begin{tabular}{lccc}
    \toprule
    \textbf{\textsc{Clauses}} & Precision & Recall & F1\\
    \midrule
    Baseline & .904 & .897 & .900\\
    \hspace{1mm} + Page Layout & .902 (-0.2\%) & .897 & .899 (-0.1\%)\\
    \hspace{1mm} + Text Placement & .908 (+0.5\%) & .897 & .902 (+0.3\%)\\
    \hspace{1mm} + Visual Grouping & .912 (+0.9\%) & .899 (+0.3\%) & .905 (+0.6\%)\\
    \hspace{1mm} + Style & .917 (+1.5\%) & \textbf{.902 (+0.6\%)} & .909 (+1.0\%)\\
    \hspace{1mm} + All Groups & \textbf{.919 (+1.7\%)} & .901 (+0.5\%) & \textbf{.910 (+1.1\%)}\\
    \midrule
    \textbf{\textsc{Sub-clauses}} & Precision & Recall & F1\\
    \midrule
    Baseline & .901 & .913 & .907\\
    \hspace{1mm} + Page Layout & .900 (-0.2\%) & .913 & .906 (-0.1\%)\\
    \hspace{1mm} + Text Placement & .901 & .913 & .907\\
    \hspace{1mm} + Visual Grouping & .904 (+0.3\%) & \textbf{.914 (+0.1\%)} & .909 (+0.2\%)\\
    \hspace{1mm} + Style & .908 (+0.7\%) & .913 & .910 (+0.4\%)\\
    \hspace{1mm} + All Groups & \textbf{.910 (+0.9\%)} & .913 & \textbf{.911 (+0.4\%)}\\
    \midrule
    \textbf{\textsc{Headers}} & Precision & Recall & F1\\
    \midrule
    Baseline & .900 & .956 & .927\\
    \hspace{1mm} + Page Layout & .840 (-6.7\%) & \textbf{.961 (+0.5\%)} & .896 (-3.3\%)\\
    \hspace{1mm} + Text Placement & .845 (-6.1\%) & .955 (-0.1\%) & .897 (-3.3\%)\\
    \hspace{1mm} + Visual Grouping & \textbf{.910 (+1.1\%)} & .958 (+0.2\%) & \textbf{.933 (+0.7\%)}\\
    \hspace{1mm} + Style & .890 (-1.1\%) & .956 & .922 (-0.6\%)\\
    \hspace{1mm} + All Groups & .858 (-4.7\%) & .960 (+0.4\%) & .906 (-2.3\%)\\
    \midrule
    \textbf{\textsc{Footers}} & Precision & Recall & F1\\
    \midrule
    Baseline & .845 & .760 & .800\\
    \hspace{1mm} + Page Layout & .877 (+3.8\%) & \textbf{.862 (+13.4\%)} & .869 (+8.6\%)\\
    \hspace{1mm} + Text Placement & .849 (+0.5\%) & .792 (+4.2\%) & .820 (+2.4\%)\\
    \hspace{1mm} + Visual Grouping & .855 (+1.2\%) & .834 (+9.7\%) & .844 (+5.5\%)\\
    \hspace{1mm} + Style & .843 (-0.2\%) & .757 (-0.4\%)& .798 (-0.3\%)\\
    \hspace{1mm} + All Groups & \textbf{.887 (+4.9\%)} & .857 (+12.8\%) & \textbf{.872 (+8.9\%)}\\
  \bottomrule
\end{tabular}
\end{table}

\subsection{Document Understanding Tasks}

Using the outputs of the section splitting process, we select relevant sections and input them into downstream models trained to extract key metadata from contracts. The four tasks we evaluate reflect four common questions lawyers need to know about their contracts:

\begin{itemize}
\item Which contracts are still active? (``Expiration Date'', framed as an entity extraction task to find the correct date in the contract text)
\item Which state or country's law governs this contract? (``Governing Law'', framed as an entity extraction task to find the correct location in the contract text)
\item Can I terminate this contract? (``Termination for Convenience'', framed as a classification task to answer yes or no correctly based on the contract text)
\item Which contracts will survive a merger or acquisition? (``Anti-Assignment'', framed as a classification task to answer yes or no correctly based on the contract text)
\end{itemize}

We evaluate against two competing approaches:
\begin{itemize}
    \item \textbf{Expert rules}: Regex-like rules written by trained annotators referencing a corpus of legal documents that does not include the CUAD contracts. For example, an annotator might determine that if the phrase ``may terminate at will'' appears in a contract, that contract's value for the ``Termination for Convenience'' attribute is ``yes''. Because of the differences between the contracts in the annotators' reference corpus and the contracts in CUAD, these rules have high precision but low recall, since they cannot generalize to unseen language in a new corpus.
    \item \textbf{DeBERTa}: The previously best-performing model on CUAD \cite{cuad2021}, a DeBERTa-xlarge model fine-tuned for the question answering task.
    We compare to the reported precision at 80\% recall for four attributes.
\end{itemize}

Results for the full end-to-end tasks are detailed in \autoref{tab:results}. Our method, which uses rich visual cues to split contracts into sections for use on downstream tasks, offers the best balance of precision and recall across the four tasks. Additionally, examples of correct and incorrect model predictions on each task are included in \autoref{appendix_model}.

\begin{table}[!htb]
  \caption{Results on End-to-End Long Document Understanding Tasks}
  \label{tab:results}
  \begin{tabular}{lccc|ccc}
    \toprule
    \multicolumn{7}{c}{\textsc{\textbf{Entity Extraction Tasks}}}\\
     & \multicolumn{3}{c}{Expiration Date} & \multicolumn{3}{c}{Governing Law}\\
    Model & P & R & F1 & P & R & F1\\
    \midrule
    Expert rules & .77 & .64 & .70 & .75 & .60 & .67\\
    DeBERTa & .86 & .80 & .83 & .97 & .80 & .88 \\
    Our model & \textbf{.87} & \textbf{.87} & \textbf{.87} & \textbf{.98} & \textbf{.98} & \textbf{.98} \\
    \midrule
    \multicolumn{7}{c}{\textsc{\textbf{Classification Tasks}}}\\
     & \multicolumn{3}{c}{Term. for Conv.} & \multicolumn{3}{c}{Anti-Assignment}\\
    Model & P & R & F1 & P & R & F1\\
    \midrule
    Expert rules & \textbf{.80} & .44 & .57 & .83 & .57 & .68\\
    DeBERTa & .37 & \textbf{.80} & .51 & .76 & .80 & .78\\
    Our model & .77 & .75 & \textbf{.76} & \textbf{.89} & \textbf{.88} & \textbf{.89}\\
    \bottomrule
  \end{tabular}
\end{table}

\section{Analysis} \label{analysis}

\subsection{Impact of Document Length} \label{impactofdoclength}

\begin{figure}[!htb]
  \centering
  \includegraphics[width=\linewidth]{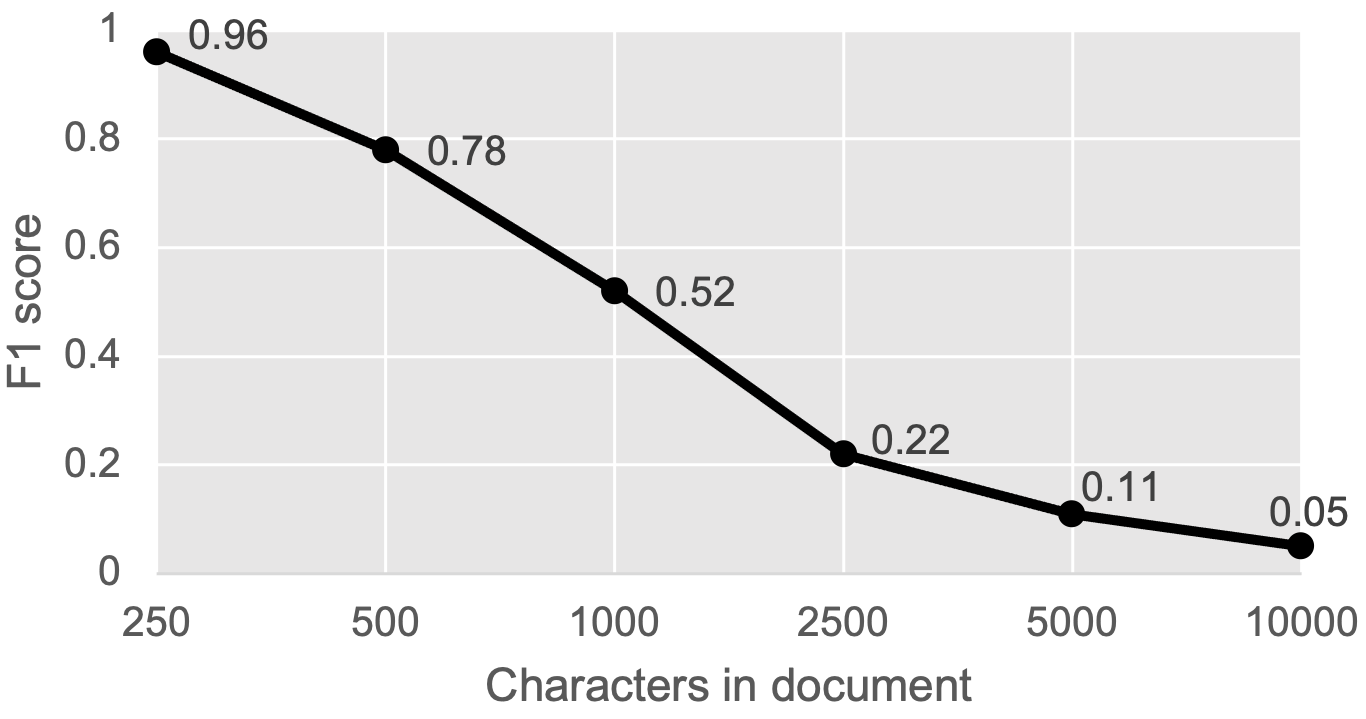}
  \caption{Longer documents are much more difficult for a CRF model extracting Governing Law from contracts of increasing size.}
  \label{fig:doclength}
  \Description{Chart showing lower performance with larger document size.}
\end{figure}

To quantify the impact of document length on the difficulty of a document understanding task, we train a CRF model to extract Governing Law from contract segments of increasing length. For each length, we draw a fixed-size window around the correct answer span in every contract. We use a random offset before and after the answer span so that answers are not always located in the same place. As shown in \autoref{fig:doclength}, as document length increases, performance sharply falls. Entity extraction on long documents is a difficult task without the benefit of visually-aware, structural features that help the model locate the answer within a document.

\subsection{Impact of OCR Features} \label{impactofocr}

Our results show strong performance on end-to-end document understanding tasks, but to what extent can we attribute that performance to the model's understanding of document structure via OCR features, as opposed to the task setup or model architecture? To quantify the impact of visual cues on one of our downstream tasks, we compare an Anti-Assignment classifier trained using sections that were obtained with and without the help of visual cues. As shown in \autoref{tab:sectionocr}, the downstream model performs even better when it has access to document structure via OCR features.

\begin{table}[!htb]
  \caption{Impact of Visual Section Splitting on Anti-Assignment Classification}
  \label{tab:sectionocr}
  \begin{tabular}{lccc}
    \toprule
    Model & P & R & F1\\
    \midrule
    Our model & .89 & .69 & .71\\
    Our model + visual cues & \textbf{.93} & \textbf{.81} & \textbf{.85}\\
    \bottomrule
  \end{tabular}
\end{table}

Finally, we provide a more detailed analysis of the performance for each section type from \autoref{tab:ocr}:

\subsubsection{Clauses}
Clauses benefit the most from style features. This is intuitive, since clauses very frequently start with a phrase that acts as a heading, set apart in bold or underlined. The visual grouping features also had a positive impact, since clauses often appear together in paragraphs of text.

\subsubsection{Sub-clauses}
Sub-clauses benefit from the same features as clauses, but it is noteworthy that the impact is lower, particularly for recall. Sub-clause detection is generally a harder problem, even for human annotators, due to deeply nested sub-clauses and a wider variety of possible ways to format them.

\subsubsection{Headers}
Headers do not see a large benefit from OCR features. In fact, page layout features hurt precision on headers, since they can be misleading when clauses and sub-clauses appear at the beginning of a page like a header. Similarly, text placement features are also less effective because the model tends to overfit to the placement of logos and auto-generated stamps like ``DocuSign Envelope Id'', which always show up in the same location. To address this weakness, we plan to add more diverse data and regularization in addition to richer features.

\subsubsection{Footers}
Footers benefit the most from OCR features. This is not surprising, since footers consistently appear at the end of the page, and are much more common than headers. Accurate footer detection is particularly useful since footers often interrupt the flow of clauses, which hurts downstream models that rely on clean section text.

\section{Conclusion}

We discuss a method for extracting valuable metadata from legal documents, which are especially challenging due to their length and their diverse structure and formatting. We show that visual cues are a key component in a successful document understanding pipeline, from section splitting to entity extraction and classification.

This work explores only a subset of the types of visual cues available. Future work might explore additional features such as font information, spacing between lines and blocks of text, superscripting, subscripting, and color. Given the impact of the visual cues we examined, we expect these features to further improve performance on document understanding tasks.

\bibliographystyle{ACM-Reference-Format}
\bibliography{main}

\appendix

\section{Model Predictions} \label{appendix_model}

We list examples of correct and incorrect model outputs for the four end-to-end tasks: Expiration Date (\autoref{tab:examples_expdate}), Governing Law (\autoref{tab:examples_govlaw}), Termination for Convenience (\autoref{tab:examples_t4c}), and Anti-Assignment (\autoref{tab:examples_aa}). ``Relevant Section Text'' refers to the output of our section splitting model, which identifies clauses and sub-clauses and then labels which sections are relevant to a given downstream task.

\begin{table*}[!p]
  \caption{Model Predictions for ``Expiration Date''}
  \label{tab:examples_expdate}
  \begin{tabular}{p{12.5cm}p{1.5cm}p{2cm}}
    \toprule
    \shortstack{Relevant Section Text} & \shortstack{Correct\\ Answer} & \shortstack{Model\\ Prediction}\\
    \midrule
    2.1 \textbf{This Agreement shall be effective from the date first above written and shall continue indefinitely until terminated by either Party in accordance with the provisions of this Agreement.} & bold text & bold text \checkmark\\
    \midrule
    The term of this Agreement shall be five (5) years. & full text & full text \checkmark\\
    \midrule
    (b)The channel \textbf{is expected to be uploaded on August 1, 2010.} & bold text & full text $\times$ \\
    \midrule
    \textbf{(i) the date on which the aggregate Required Capital Contributions paid by the Investor, and received by the Borrowers, in accordance with the terms of this Agreement equal \$5,000,000; (ii) the Release Date (as defined in Section 19) or (iii) payment in full, in cash, of all Obligations and the termination of the Financing Agreement; provided, however, that this Agreement shall continue to be effective, or be reinstated, as the case may be, if at any time the aggregate Required Capital Contributions} paid by the Investor, and received by the Borrowers, in accordance with the terms of this Agreement is less than \$5,000,000 and any payment, or any part thereof, on account of any of the Obligations is invalidated, declared to be fraudulent or preferential, set aside, rescinded or must otherwise be restored or returned by the Agent or the Lenders upon the insolvency, bankruptcy, liquidation, dissolution or reorganization of any Borrower or upon or as a result of the appointment of a receiver, intervenor or conservator of, or trustee or similar officer for any Borrower, or any substantial part of this property, or otherwise, all as though such payment had not been made. & full text & bold text $\times$ \\
    \midrule
    17. TERM AND TERMINATION (a) \textbf{This Agreement and the license granted under this Agreement shall remain in effect perpetually as long as fees are paid by Sparkling in accordance with the Fee Schedule and the Agreement is not otherwise terminated in accordance with this Section.} & bold text & bold text \checkmark\\
    \bottomrule
  \end{tabular}

\bigskip
\bigskip

  \caption{Model Predictions for ``Governing Law''}
  \label{tab:examples_govlaw}
  \begin{tabular}{p{12.5cm}p{1.5cm}p{2cm}}
    \toprule
    \shortstack{Relevant Section Text} & \shortstack{Correct\\ Answer} & \shortstack{Model\\ Prediction}\\
    \midrule
    14. Governing Law. \textbf{This Agreement shall be construed and interpreted in accordance with the laws of the State of Ohio, without recourse to any principles of law governing conflicts of law, which might otherwise be applicable.} & bold text & bold text \checkmark\\
    \midrule
    This Agreement and performance under this Agreement shall be governed by the laws of the United States of America and of the Commonwealth of Pennsylvania as applied to agreements entered into and to be performed entirely within Pennsylvania between Pennsylvania residents, excluding its conflicts of law provisions. & full text & full text \checkmark\\
    \midrule
    12.2 \textbf{This Agreement and all matters pertaining hereto shall be governed by and construed under the laws of the State of Louisiana, except to the extent} that the conflict of law rules of said state would require that the laws of another state would govern its validity, construction, or interpretation. & full text & bold text $\times$ \\
    \midrule
    This Agreement shall be governed by the laws of the Province of Ontario and the federal laws of Canada applicable therein. & full text & full text \checkmark\\
    \midrule
    Without reference to choice or conflict of law principles, this Agreement shall be governed by and construed in accordance with the laws of the State of California, USA. & full text & full text \checkmark\\
    \midrule
    (a) \textbf{This Agreement and all Actions (whether in contract or tort) that may be based upon, arise out of or relate to this Agreement or the negotiation, execution, or performance hereof or thereof shall be governed by and construed in accordance with the Law of the State of Delaware, without regard to any Laws or principles thereof that would result in the application of the Laws of any other jurisdiction.} & bold text & bold text \checkmark\\
    \bottomrule
  \end{tabular}
\end{table*}

\begin{table*}[!p]
  \caption{Model Predictions for ``Termination for Convenience''}
  \label{tab:examples_t4c}
  \begin{tabular}{p{13.5cm}p{1cm}p{1.5cm}}
    \toprule
    \shortstack{Relevant Section Text} & \shortstack{Correct\\ Answer} & \shortstack{Model\\ Prediction}\\
    \midrule
    8.2 Termination. Either Party may terminate this Agreement for the material breach or default of any of the terms or conditions of this Agreement by the other Party upon thirty (30) days' written notice and the opportunity to cure during such notice period; and such termination shall be in addition to any other remedies that it may have at law or in equity. Additionally, LBIO may terminate this Agreement if MD Anderson is declared insolvent or enters into liquidation or has a receiver or an administrator appointed over all or any part of its assets or ceases or threatens to cease to carry on business, or a resolution is passed or a petition presented to any court for the winding up of the Party or for the granting of an administration order in respect of MD Anderson, or any proceedings are commenced relating to the insolvency or possible insolvency of MD Anderson. & No & Yes $\times$ \\
    \midrule
    7.2. This Agreement may be terminated by either party with cause upon thirty (30) days written notice. Upon Marketing Affiliate's default in payment or other breach of this Agreement, Equidata may terminate this Agreement without notice to Marketing Affiliate. Upon termination for any reason, Equidata reserves the right to deactivate Marketing Affiliate's access to the services including the Equidata Web Site. Termination does not release Marketing Affiliate from paying all amounts owed to Equidata. & No & Yes $\times$ \\
    \midrule
    8 1 TERM OF AGREEMENT. This Agreement shall continue in force for a term of twelve (12) months from the Effective Date, unless terminated earlier under the provisions of this Article 8 (the ``Term''); PROVIDED that TouchStar shall have the right to terminate this Agreement at any time after the Effective Date upon not less than fifteen (15) days' prior written notice to Reseller. Prior to the end of the Term, each of TouchStar and Reseller may notify the other if it desires to negotiate a further agreement by written request received at least ninety (90) days in advance of the termination of this Agreement. If both parties desire to negotiate a further agreement, they may consider the terms of this Agreement in coming to an understanding. Nothing in this Agreement shall be construed to obligate either party to renew or extend the term of this Agreement. Renewals for additional terms, if any, shall not cause this Agreement to be construed as an agreement of indefinite duration. & Yes & No $\times$ \\
    \midrule
    18.1 The Company may terminate the Executive's employment under this Agreement with immediate effect without notice and with no liability to make any further payment to the Executive (other than in respect of amounts accrued at the Termination Date) if in the reasonable opinion of the Company the Executive: & Yes & No $\times$ \\
    \midrule
    8.2. Termination for Cause. Either party may terminate this Agreement immediately upon written notice to the other party in the event any material breach of a material term of this Agreement by such other party that remains uncured 30 days in the case of a breach of a payment obligation, or 45 days for all other breaches, after notice of such breach was received by such other party; provided, however that if such breach is not reasonably capable of cure within the applicable cure period, the breaching party shall have an additional 180 days to cure such breach so long as the cure is commenced within the applicable cure period and thereafter is diligently prosecuted to completion as soon as possible. & No & Yes $\times$ \\
    \midrule
    4.2 The term of this Agreement is for a period of five (5) years (the ``Term'') commencing on the Effective Date and, unless terminated earlier in accordance with the termination provisions of this Agreement, ending on January 31, 2025. & Yes & No $\times$ \\
    \midrule
    Section 2 — Term. This Agreement shall commence as of the Effective Date and shall continue in full force and effect for an initial term of three (3) years from the Promotion Commencement Date, divided into three one-year periods. Unless terminated in accordance with the provisions of Section 18, this Agreement shall automatically renew for each subsequent one-year term. & No & No \checkmark\\
    \midrule
    TERM AND TERMINATION. A. By either party as a result of default by the other party under this Agreement and failure to cure said default within thirty (30) days after notice of said default is given. & No & Yes $\times$ \\
    \midrule
    Terms and Termination: The term of this agreement will begin on April 1, 2018 and continue until April 30, 2018 at 11:59pm. & No & No \checkmark\\
    \bottomrule
  \end{tabular}
\end{table*}

\begin{table*}[!p]
  \caption{Model Predictions for ``Anti-Assignment''}
  \label{tab:examples_aa}
  \begin{tabular}{p{13.5cm}p{1cm}p{1.5cm}}
    \toprule
    \shortstack{Relevant Section Text} & \shortstack{Correct\\ Answer} & \shortstack{Model\\ Prediction}\\
    \midrule
    VII. ASSIGNMENT. Neither this Agreement nor any rights or obligations or licenses hereunder may be assigned, pledged, transferred or encumbered by either party without the express prior written approval of the other party, except that either HEMISPHERX or SCIEN may assign this Agreement to any successor by merger or sale of substantially all of its business or assets to which this Agreement pertains, without any such consent. Any assignment in violation hereof is void. & Yes & Yes \checkmark\\
    \midrule
    ASSIGNMENT: NOW, THEREFORE, for good and valuable consideration, the receipt and adequacy of which are hereby acknowledged, Seller does hereby transfer, sell, assign, convey and deliver to Backhaul all right, title and interest in, to and under the Assigned Intellectual Property, including, without limitation, the Trademarks and Patents set forth on Schedules A and B hereof, respectively, and all goodwill of the Purchased Business associated therewith. Seller hereby covenants and agrees, that from time to time forthwith upon the reasonable written request of Backhaul or Buyer, that Seller will, at Backhaul's cost and expense, do, execute, acknowledge and deliver or cause to be done, executed, acknowledged and delivered, each and all of such further acts, deeds, assignments, transfers, conveyances and assurances as may reasonably be required by Backhaul or Buyer in order to transfer, assign, convey and deliver unto and vest in Backhaul title to all right, title and interest of Seller in, to and under the Assigned Intellectual Property. & No & Yes $\times$ \\
    \midrule
    (no relevant section found) & Yes & No $\times$ \\
    \midrule
    4.4 Assignments and Transfers by Seller Trusts. The provisions of this OMA shall be binding upon and inure to the benefit of the Seller Trusts and their respective successors and assigns. A Seller Trust may transfer or assign, in whole or from time to time in part, to one or more liquidating trusts its rights hereunder in connection with the transfer or resale of Stock held by such Seller Trust, provided that such Seller Trust complies with all laws applicable thereto and provides written notice of assignment to GWG promptly after such assignment is effected, and provided further that such liquidating trust and each beneficiary thereof executes a joinder to this OMA effective as of the date of such assignment or transfer. & No & Yes $\times$ \\
    \midrule
    133 Assignment. Neither Party shall assign this Development Agreement or the obligations contained herein without the express written consent of the other Party. & Yes & Yes \checkmark\\
    \midrule
    15.7 Successors and Assigns. This Agreement shall inure to the benefit of and be binding upon the Parties and their respective successors and assigns, including, but not limited to, any chapter 11 or chapter 7 trustee; provided, however, that this Agreement may not be assigned by any of the Parties without the prior written consent of the other, provided further that notwithstanding the foregoing, GA and Tiger may each collaterally assign this Agreement and their rights thereunder to their respective lenders. & Yes & Yes \checkmark\\
    \midrule
    9.2 Assignment 51 & Yes & Yes \checkmark\\
    \bottomrule
  \end{tabular}
\end{table*}

\section{Dataset} \label{app:dataset}

\subsection{Details on CUAD}

\autoref{tab:appdata} lists detailed statistics on the contracts in CUAD.

CUAD provides both PDF and text versions of contracts. For our experiments, we use the PDF versions so that we can extract OCR metadata from the formatting and structure of the contracts.

\begin{table*}
  \caption{Statistics on the Contract Understanding Atticus Dataset (CUAD)}
  \label{tab:appdata}
  \begin{tabular}{lr}
    \toprule
    Total documents & 510 \\
    Documents in train set (80\%) & 408 \\
    Documents in dev/test sets (10\%/10\%) & 51 \\
    Average characters per contract & 52,563 \\
    Characters in shortest contract & 645 \\
    Characters in longest contract & 338,211 \\
    Average words per contract & 9,594 \\
    Words in shortest contract & 109 \\
    Words in longest contract & 103,923 \\
  \bottomrule
\end{tabular}
\end{table*}

\subsection{Attributes Chosen from CUAD}

While CUAD includes labels for the two most useful contract metadata attributes, Document Name and Parties, we choose not to evaluate using these labels because both of these attributes occur multiple times throughout each document, and the label offsets in CUAD do not consistently come from the same location in the document. For example, some Document Name labels are selected from an Appendix rather than the beginning of the document, making evaluation on these labels misleading.

\subsection{Test Set}

To facilitate future work on CUAD, we list the filenames of the 51 documents in our test set in \autoref{tab:filenames}.

\begin{table*}
  \caption{Filenames of documents from CUAD in our test set}
  \label{tab:filenames}
  \begin{tabular}{l}
  \toprule
Monsanto Company - SECOND A\&R EXCLUSIVE AGENCY AND MARKETING AGREEMENT .PDF\\
IdeanomicsInc\_20151124\_8-K\_EX-10.2\_9354744\_EX-10.2\_Content License Agreement.pdf\\
REGANHOLDINGCORP\_03\_31\_2008-EX-10-LICENSE AND HOSTING AGREEMENT.PDF\\
GridironBionutrientsInc\_20171206\_8-K\_EX-10.1\_10972555\_EX-10.1\_Endorsement Agreement.pdf\\
BLACKBOXSTOCKSINC\_08\_05\_2014-EX-10.1-DISTRIBUTOR AGREEMENT.PDF\\
OLDAPIWIND-DOWNLTD\_01\_08\_2016-EX-1.3-AGENCY AGREEMENT2.pdf\\
ClickstreamCorp\_20200330\_1-A\_EX1A-6\_MAT CTRCT\_12089935\_EX1A-6\_MAT CTRCT\_Development Agreement.pdf\\
NYLIACVARIABLEANNUITYSEPARATEACCOUNTIII\_04\_10\_2020-EX-99.8.KK-SERVICE AGREEMENT.PDF\\
Columbia Laboratories (Bermuda)Ltd. - AMEND NO. 2 TO MANUFACTURING AND SUPPLY AGREEMENT.PDF\\
MERITLIFEINSURANCECO\_06\_19\_2020-EX-10.(XIV)-MASTER SERVICES AGREEMENT.PDF\\
NATIONALPROCESSINGINC\_07\_18\_1996-EX-10.4-SPONSORSHIP AGREEMENT.PDF\\
MJBIOTECHINC\_12\_06\_2018-EX-99.01-JOINT VENTURE AGREEMENT.PDF\\
KUBIENTINC\_07\_02\_2020-EX-10.14-MASTER SERVICES AGREEMENT\_Part2.pdf\\
ArcGroupInc\_20171211\_8-K\_EX-10.1\_10976103\_EX-10.1\_Sponsorship Agreement.pdf\\
WORLDWIDESTRATEGIESINC\_11\_02\_2005-EX-10-RESELLER AGREEMENT.PDF\\
SPOKHOLDINGSINC\_06\_19\_2020-EX-10.1-COOPERATION AGREEMENT.PDF\\
LegacyEducationAllianceInc\_20200330\_10-K\_EX-10.18\_12090678\_EX-10.18\_Development Agreement.pdf\\
HEALTHGATEDATACORP\_11\_24\_1999-EX-10.1-HOSTING AND MANAGEMENT AGREEMENT - Escrow Agreement.pdf\\
LECLANCHE S.A. - JOINT DEVELOPMENT AND MARKETING AGREEMENT.PDF\\
THERAVANCEBIOPHARMA,INC\_05\_08\_2020-EX-10.2-SERVICE AGREEMENT.PDF\\
HUBEIMINKANGPHARMACEUTICALLTD\_09\_19\_2006-EX-10.1-OUTSOURCING AGREEMENT.PDF\\
GSITECHNOLOGYINC\_11\_16\_2009-EX-10.2-INTELLECTUAL PROPERTY AGREEMENT between SONY ELECTRONICS INC.\\
\hspace{3mm}and GSI TECHNOLOGY, INC..PDF\\
SECURIANFUNDSTRUST\_05\_01\_2012-EX-99.28.H.9-NET INVESTMENT INCOME MAINTENANCE AGREEMENT.PDF\\
HarpoonTherapeuticsInc\_20200312\_10-K\_EX-10.18\_12051356\_EX-10.18\_Development Agreement.PDF\\
InnerscopeHearingTechnologiesInc\_20181109\_8-K\_EX-10.6\_11419704\_EX-10.6\_Distributor Agreement.pdf\\
SoupmanInc\_20150814\_8-K\_EX-10.1\_9230148\_EX-10.1\_Franchise Agreement1.pdf\\
EdietsComInc\_20001030\_10QSB\_EX-10.4\_2606646\_EX-10.4\_Co-Branding Agreement.pdf\\
FerroglobePlc\_20150624\_F-4A\_EX-10.20\_9154746\_EX-10.20\_Outsourcing Agreement.pdf\\
BIOCEPTINC\_08\_19\_2013-EX-10-COLLABORATION AGREEMENT.PDF\\
HertzGroupRealtyTrustInc\_20190920\_S-11A\_EX-10.8\_11816941\_EX-10.8\_Trademark License Agreement.pdf\\
MEDALISTDIVERSIFIEDREIT,INC\_05\_18\_2020-EX-10.1-CONSULTING AGREEMENT.PDF\\
BIOFRONTERAAG\_04\_29\_2019-EX-4.17-SUPPLYAGREEMENT.PDF\\
MSCIINC\_02\_28\_2008-EX-10.10-.PDF\\
FEDERATEDGOVERNMENTINCOMESECURITIESINC\_04\_28\_2020-EX-99.SERV AGREE-SERVICES AGREEMENT\_POWEROF.pdf\\
CUROGROUPHOLDINGSCORP\_05\_04\_2020-EX-10.3-SERVICING AGREEMENT.PDF\\
NOVOINTEGRATEDSCIENCES,INC\_12\_23\_2019-EX-10.1-JOINT VENTURE AGREEMENT.PDF\\
TALCOTTRESOLUTIONLIFEINSURANCECO-SEPARATEACCOUNTTWELVE\_04\_30\_2020-EX-99.8(L)-SERVICE AGREEMENT.PDF\\
CORALGOLDRESOURCES,LTD\_05\_28\_2020-EX-4.1-CONSULTING AGREEMENT.PDF\\
ChinaRealEstateInformationCorp\_20090929\_F-1\_EX-10.32\_4771615\_EX-10.32\_Content License Agreement.pdf\\
GLOBALTECHNOLOGIESLTD\_06\_08\_2020-EX-10.16-CONSULTING AGREEMENT.PDF\\
SalesforcecomInc\_20171122\_10-Q\_EX-10.1\_10961535\_EX-10.1\_Reseller Agreement.pdf\\
MANUFACTURERSSERVICESLTD\_06\_05\_2000-EX-10.14-OUTSOURCING AGREEMENT.PDF\\
ONEMAINHOLDINGS,INC\_02\_20\_2020-EX-99.D-JOINT FILING AGREEMENT.PDF\\
BerkshireHillsBancorpInc\_20120809\_10-Q\_EX-10.16\_7708169\_EX-10.16\_Endorsement Agreement.pdf\\
VAXCYTE,INC\_05\_22\_2020-EX-10.19-SUPPLY AGREEMENT.PDF\\
SEASPINEHOLDINGSCORP\_10\_10\_2018-EX-10.1-SUPPLY AGREEMENT.PDF\\
WaterNowInc\_20191120\_10-Q\_EX-10.12\_11900227\_EX-10.12\_Distributor Agreement.pdf\\
ElPolloLocoHoldingsInc\_20200306\_10-K\_EX-10.16\_12041700\_EX-10.16\_Development Agreement.pdf\\
TubeMediaCorp\_20060310\_8-K\_EX-10.1\_513921\_EX-10.1\_Affiliate Agreement.pdf\\
IOVANCEBIOTHERAPEUTICS,INC\_08\_03\_2017-EX-10.1-STRATEGIC ALLIANCE AGREEMENT.PDF\\
VertexEnergyInc\_20200113\_8-K\_EX-10.1\_11943624\_EX-10.1\_Marketing Agreement.pdf\\
  \bottomrule
  \end{tabular}
\end{table*}

\end{document}